\documentclass[letterpaper, 10 pt, conference]{ieeeconf}

\IEEEoverridecommandlockouts 
\usepackage{amsmath,amsfonts}
\usepackage{algorithmic}
\usepackage{algorithm}
\usepackage{array}
\usepackage[caption=false,font=normalsize,labelfont=sf,textfont=sf]{subfig}
\usepackage{textcomp}
\usepackage{stfloats}
\usepackage{url}
\usepackage{verbatim}
\usepackage{graphicx}
\usepackage{cite}
\usepackage{todonotes}
\usepackage{amssymb}
\usepackage{booktabs}
\usepackage{makecell}
\usepackage{balance}

\newif\ifanonymous

\ifanonymous
    \author{Anonymous Authors}
\else
    \author{Vladyslav Havriutkin$^1$, Fabio Elnecave Xavier$^2$, Marine Petriaux$^2$, Frank Dellaert$^1$, Maegan Tucker$^1$
    \thanks{$^1$ Authors are with Georgia Institute of Technology, Atlanta, U.S.}
    \thanks{$^2$ Authors are with Wandercraft, Paris, France.}
    \thanks{\texttt{mtucker@gatech.edu}}
    }
\fi

\begin{document}

\title{Proprioceptive State Estimation for \\ Lower-Limb Exoskeleton via Factor Graphs}



\maketitle

\begin{abstract}
Accurate state estimation for exoskeletons is challenging due to significant load-induced mechanical deformations, multiple sensor modalities, and unpredictable contact slippage. Compared to traditional filtering techniques which suffer from marginalization and linearization errors, and purely data-driven methods which lack formal safety guarantees, factor graphs offer a robust alternative by framing estimation as a nonlinear smoothing problem. In this work, we propose a factor-graph-based state estimator for the Atalante exoskeleton that fuses measurements from twelve joint encoders, five inertial measurement units (IMUs), and eight vertical force sensors. To account for unmodeled structural bending and foot slippage, we introduce a probabilistic contact formulation that dynamically adjusts factor covariances using real-time force telemetry. Experimental validation during dynamic walking, benchmarked against motion capture ground truth, demonstrates that the proposed estimator provides accurate estimates of robot's pose in a world frame, outperforming naive EKF approaches.
\end{abstract}


\section{Introduction}
Lower-limb exoskeletons for people with motor complete paraplegia are complex multi-sensor systems that can enable crutch-less walking in the absense of voluntary lower-limb motor function \cite{harib2018feedback, fabio-ekf-imu-atalante, lower-limb-exo, teleop-of-exo}. While it is common for biped robots to suffer mechanical deformations, exoskeletons suffer from significant mechanical bending due to the additional weight of the human user, which subjects the structure to high-magnitude loads \cite{benallegue2015estimation, vigne2019state}. This leads to multi-centimeter kinematic displacements at the feet \cite{MOVIE}. To satisfy strict safety requirements in the presence of these deformations, modern exoskeletons utilize redundant sensor architectures. Consequently, the integration of multiple inertial measurement units (IMUs), foot contact sensors, and joint encoders presents a challenging sensor fusion problem. Accurate state estimation of the exoskeleton's kinematics is crucial for executing stable control and ensuring safe human-robot interaction in unpredictable environments. Existing approaches range from classical filtering to data-based learning methods \cite{vigne2019state, MOVIE, fabio-ekf-imu-atalante, buchanan2022learning}.

Filtering techniques, such as the Extended Kalman Filter (EKF), dominate the field \cite{EKF-Est} \cite{Contact-Aided-InEKF}, but have fundamental limitations. Filters marginalize past variables, preventing the re-evaluation of past states. As the number of available sensors increases, the EKF covariance matrix becomes large and dense, introducing computational bottlenecks. Furthermore, filtering restricts the handling of nonlinear dynamics to a single-point linearization at the current timestep, which leads to divergence during swing-foot impact events. While recent EKF architectures have successfully integrated multiple IMUs on the Atalante exoskeleton \cite{fabio-ekf-imu-atalante}, filtering frameworks remain difficult to scale and lose accuracy due to marginalization.

Learning-based methods have demonstrated efficacy in data-driven state estimation \cite{ML-in-legged-robot, ML-learning-for-quadruped}, but remain difficult to use in safety-critical applications in wearable robotics. Purely data-driven end-to-end pipelines currently lack formal safety and observability guarantees required for robust impact analysis in human-in-the-loop systems. Additionally, learned models struggle to generalize outside of their training distribution, which is particularly relevant for exoskeleton systems with varying human users. 

Factor graphs, which express a joint objective as a product of local factors over small subsets of variables and thereby expose the sparsity that makes large-scale inference tractable \cite{kschischang2001factor,dellaert2021factor}, have been used in robotics for state estimation \cite{forster2016manifold, fg-for-perception, fg-for-estimation-in-legged-robotics}, planning \cite{dong2016motion, mukadam2018continuous, fg-kinodynamics-planning} and control \cite{fg-mpc-uav, hoffmann2017linear, yang2021equality, bazzana2022handling, king2022simultaneous}. Of particular interest is the work of \cite{fg-for-estimation-in-legged-robotics} in which factor graphs demonstrated strong performance towards state estimation for quadrupeds. Nevertheless, factor graphs application to bipedal exoskeletons, subject to severe load-induced deformations, remains largely unexplored.

In this work, we formulate exoskeleton state estimation as a factor graph optimization problem. By retaining a localized temporal history of states and measurements, this architecture frames state estimation as a nonlinear smoothing problem. This allows for iterative re-linearization, improving estimator accuracy over EKF approaches (see Table \ref{tab:results}). Furthermore, factor graphs exploit the inherent sparsity of the sensor constraints to efficiently scale with high-dimensional data. Unlike black-box solutions, factor graphs have the expressive ability to represent large-scale problems in a natural and human-readable graphical language. Additionally, we utilize force sensor telemetry to dynamically adjust covariances of contact factors. We demonstrate that this allows the optimizer to preserve an accurate global yaw estimation.
The primary contributions of this work are:
\begin{enumerate}
    \item The design and implementation of a factor-graph-based state estimator for the Atalante lower-limb exoskeleton, fusing kinematic constraints with joint encoders, five distributed IMUs, and eight force sensors.

    \item Experimental validation of the proposed estimator on the Atalante lower-limb exoskeleton during dynamic walking, benchmarked against ground truth estimates obtained from motion capture and compared with EKF baselines.
\end{enumerate}
In addition, upon acceptance, source code and an example will be made available as an open-source resource.


\section{Atalante Exoskeleton}
\label{sec:system-model}

\begin{figure}[tb]
\vspace*{3mm} 
\centering
\includegraphics[width=\columnwidth]{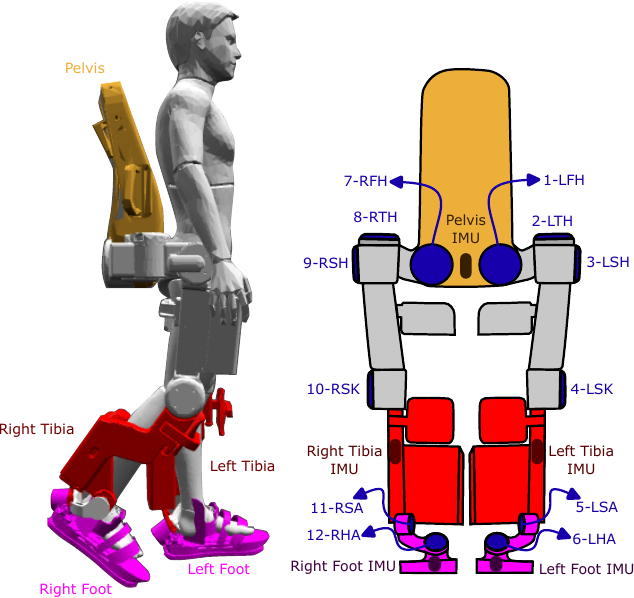}
\caption{Left) Side view of the Atalante exoskeleton with the five links containing an IMU sensor highlighted: the pelvis link is shown in yellow, the tibia links are shown in red, and the feet are shown in purple. Right) Front view of Atalante with each of the twelve actuated DOFs illustrated and numbered in blue: Left frontal hip (LFH), left transverse hip (LTH), left sagittal hip (LSH), left sagittal knee (LSK), left sagittal ankle (LSA), left henke ankle (LHA), and the respective joints on the right side (RFH,RTH,RSK,RSA,RHA).}
\label{fig:exo_frames}
\end{figure}

\noindent
The Atalante exoskeleton, illustrated in Fig. \ref{fig:exo_frames}, is modeled as a floating-base kinematic tree with $18$ degrees of freedom (DoF). The configuration space comprises $6$ unactuated spatial DoF representing the global pose of the pelvis, and $12$ actuated revolute joints $\mathbf{q} \in \mathbb{R}^{12}$ ($6$ per leg).

To formulate the estimation problem, we define a fixed inertial world frame, $\mathcal{W}$, whose origin is established at ground level during initialization with the $z$-axis opposing gravity and following Forward-Left-Up (FLU) convention. Each joint in the kinematic chain is associated with a local link frame $L_i$. The rigid transformations between consecutive link frames are governed by the joint configurations $\mathbf{q}$ via the system's forward kinematics. 
Additionally, each link possesses a center of mass (CoM) frame $C_i$, related to its corresponding link frame by a constant static transformation. 

We assume that the ground is flat and at least one foot is in contact at all times. We denote the active contact frame as $\mathcal{C}$, corresponding to the link frame of the planted foot. In double support phases, $\mathcal{C}$ is assigned to the foot registering the highest normal force. Finally, we define a local frame $\mathcal{L}$. This frame is translationally coincident with the contact frame $\mathcal{C}$ projected to the floor, but its yaw orientation is aligned with the robot's forward axis in the world frame. 

The Atalante hardware includes 12 joint encoders, 8 vertical force sensors (4 per foot), and 6 inertial measurement units (IMUs), all operating at 1000 Hz. Because our focus is on estimating the exoskeleton's intrinsic state, we exclude a user-mounted Thorax IMU from our formulation. The remaining five IMUs are rigidly attached within the pelvis link of the exoskeleton, within the left/right tibia link, and within the left/right foot links. 

The primary objective of the estimator is to track the trajectory of the pelvis and feet link poses, $T_k^P, T_k^{LF},T_k^{RF} \in SE(3)$, with respect to both the global world frame $\mathcal{W}$ and the local frame $\mathcal{L}$, where $k \in \mathbb{N}$ is the discrete time index along the trajectory.

\section{Methodology}
\label{sec:method}
\noindent
The core of our estimator is a factor graph. Factor graphs contain two types of nodes: factors and variables. Factors represent objectives and constraints that are optimized for the variables (poses, twists and biases). 

A fundamental philosophy of our approach is the use of dynamic factor covariances to handle unmodeled system dynamics, specifically mechanical deformations and contact slippage. As previously mentioned, the Atalante exoskeleton ends significantly under high-magnitude loads, which conflicts with standard kinematic models that assume perfectly rigid links. Rather than explicitly modeling these complex structural deformations, we absorb them probabilistically into the factors' noise models. This allows the optimizer to rely less on the rigid kinematic model during impacts.

We first describe variables and factors that we utilize, and then the overall graph topology used in the estimator.

\subsection{Kinematic Factors}
\noindent
To enforce rigid-body and velocity constraints, we introduce compound two-way kinematic factors. Instead of treating individual joint measurements as independent variables, these factors encapsulate the forward kinematics between specified links given joint measurements. 

The \emph{forward kinematics factor} constrains the relative poses between selected links within a kinematic chain. Consider the chain from the Right Tibia to the Right Foot (see Fig.~\ref{fig:exo_frames}), connected by two measured joints, $q_{11}$ and $q_{12}$. This factor minimizes the manifold distance between the expected relative pose ${}^{RT}T_{RF}(q_{11}, q_{12})$, derived from the joint measurements, and the prediction ${}^{w}T_{RT}^{-1} {}^{w}T_{RF}$ derived from the unknown pose variables ${}^{w}T_{RT}$ and ${}^{w}T_{RF} \in SE(3)$. The error function for this specific chain is formulated compactly as:
\begin{equation}
\label{eq:fk-factor-specific}
\begin{split}
\mathbf{e}_{\text{fk}}({}^{w}T_{RT}, {}^{w}T_{RF}&; q_{11}, q_{12}) = \\
&\log \left( {}^{RT}T_{RF}(q_{11}, q_{12})^{-1} {}^{w}T_{RT}^{-1} {}^{w}T_{RF} \right)^{\vee}.
\end{split}
\end{equation}

For any start link $s$ and end link $e$ separated by a vector of measured joint angles $\mathbf{q}$, this generalizes to:
\begin{equation}
\label{eq:fk-factor}
\mathbf{e}_{\text{fk}}({}^{w}T_{s}, {}^{w}T_{e}; \mathbf{q}) = \log \left( {}^{s}T_{e}(\mathbf{q})^{-1} {}^{w}T_{s}^{-1} {}^{w}T_{e} \right)^{\vee},
\end{equation}
where $\log: SE(3) \to \mathfrak{se}(3)$ is the logarithmic map, and $\vee: \mathfrak{se}(3) \to \mathbb{R}^6$ is the vee operator, which together map the manifold error into the vector tangent space.

Similarly, the \emph{twist factor} enforces first-order kinematic constraints on relative velocities between two selected links in a chain. Continuing with the previous example, the Right Tibia and Right Foot are connected through joints $q_{11}$ and $q_{12}$, but we now account for their corresponding joint velocities, $\dot{q}_{11}$ and $\dot{q}_{12}$. The error function enforcing the kinematic relationship between their body twists $\xi_{RT}, \xi_{RF} \in \mathbb{R}^{6}$ is expressed as:
\begin{equation} 
\label{eq:twist-factor-specific} 
\begin{split} 
\mathbf{e}_{\text{twist}}&(\xi_{RT}, \xi_{RF}; \mathbf{q}, \dot{\mathbf{q}}) = \xi_{RF} - \Bigg( \mathbf{Ad}_{{}^{RF}T_{RT}(\mathbf{q})} \xi_{RT} \\ & + \mathbf{Ad}_{{}^{RF}T_{L_{11}}(\mathbf{q})} \mathcal{A}_{11} \dot{q}_{11} + \mathbf{Ad}_{{}^{RF}T_{L_{12}}(\mathbf{q})} \mathcal{A}_{12} \dot{q}_{12} \Bigg),
\end{split} 
\end{equation}
where $\mathcal{A}_i$ are the individual joint screw axes and $\mathbf{Ad}_{{}^{e}T_{s}(\mathbf{q})} \in \mathbb{R}^{6 \times 6}$ is the adjoint matrix mapping spatial velocities from the start frame to the end frame.

This formulation generalizes for body twists $\xi_{s}, \xi_{e}$ across any link chain given corresponding joint measurements $\mathbf{q}, \dot{\mathbf{q}}$:
\begin{equation}
\label{eq:twist-factor}
\mathbf{e}_{\text{twist}}(\xi_{s}, \xi_{e}; \mathbf{q}, \dot{\mathbf{q}}) = \xi_{e} - \left( \mathbf{Ad}_{{}^{e}T_{s}(\mathbf{q})} \xi_{s} + \xi_{\text{rel}}(\mathbf{q}, \dot{\mathbf{q}}) \right).
\end{equation}
The term $\xi_{\text{rel}}(\mathbf{q}, \dot{\mathbf{q}}) \in \mathbb{R}^{6}$ represents the relative twist induced at the end link by the internal joint velocities, defined by the sum of the individual joint screw axes $\mathcal{A}_i$ transformed into the end-effector frame:
\begin{equation} 
\label{eq:relative-twist} 
\xi_{\text{rel}}(\mathbf{q}, \dot{\mathbf{q}}) = \sum_{i} \mathbf{Ad}_{{}^{e}T_{L_i}(\mathbf{q})} \mathcal{A}_i \dot{q}_i,
\end{equation}
where $L_i$ is a link frame associated with joint $i$.

\subsection{IMU Factors}
\noindent
To incorporate IMU measurements, the \emph{combined IMU factor} \cite{gtsam-imu-1, gtsam-imu-2, gtsam-imu-3-preintegration} from GTSAM \cite{gtsam} is used. The combined IMU factor is a six-way factor connecting pose and twist variables of a link with IMU bias variables at two consecutive slices. The overall error of this factor consists of four components:
\begin{equation}
\label{eq:imu-factor}
\mathbf{e}_{\text{IMU}} = 
\begin{bmatrix}
\mathbf{e}_{\Delta R} \\
\mathbf{e}_{\Delta v} \\
\mathbf{e}_{\Delta p} \\
\mathbf{e}_{\Delta b} 
\end{bmatrix} \in \mathbb{R}^{15},
\end{equation}
with $\mathbf{e}_{\Delta R}\in \mathbb{R}^3$ denoting the rotational error, $\mathbf{e}_{\Delta v}\in \mathbb{R}^3$ the velocity error, $\mathbf{e}_{\Delta p}\in \mathbb{R}^3 $ the positional error, and $\mathbf{e}_{\Delta b}\in \mathbb{R}^6$ the bias evolution error.

Specifically, the rotational error is represented as: 
\begin{equation}
\label{eq:imu-factor-rot}
\mathbf{e}_{\Delta R} = \log \left( \left[ \Delta \tilde{R}_{ij} \cdot \exp \left( \frac{\partial \Delta R}{\partial \mathbf{b}_g} \delta \mathbf{b}_{g,i} \right) \right]^\top \cdot R_i^\top R_j \right)^\vee,
\end{equation}
where $\Delta \tilde{R}_{ij} \in SO(3)$ is the preintegrated relative rotation measurement; $R_i, R_j \in SO(3)$ are the orientation matrices of the link's CoM frame in the world frame at consecutive times $i$ and $j$; $\frac{\partial \Delta R}{\partial \mathbf{b}_g} \in \mathbb{R}^{3 \times 3}$ is the Jacobian of the preintegrated rotation with respect to the gyroscope bias; $\delta \mathbf{b}_{g,i} \in \mathbb{R}^3$ is the gyroscope bias deviation from the linearization point at time $i$; $\exp: \mathbb{R}^3 \to SO(3)$ is the matrix exponential mapping a tangent space vector to the manifold; and $\log: SO(3) \to \mathbb{R}^3$ is the logarithmic map, with $\vee$ extracting the tangent space vector from the $SO(3)$ manifold error.

The velocity error is represented as:
\begin{equation}
\label{eq:imu-factor-vel}
\begin{split}
\mathbf{e}_{\Delta v} ={}& R_i^\top (v_j - v_i - \mathbf{g}\Delta t) \\
&- \left( \Delta \tilde{v}_{ij} + \frac{\partial \Delta v}{\partial \mathbf{b}_g} \delta \mathbf{b}_{g,i} + \frac{\partial \Delta v}{\partial \mathbf{b}_a} \delta \mathbf{b}_{a,i} \right),
\end{split}
\end{equation}
where $R_i \in SO(3)$ is the (estimated) rotational component of the link's pose variable at time $i$; $v_i, v_j \in \mathbb{R}^3$ are the spatial velocities in the world frame at times $i$ and $j$; $\mathbf{g} \in \mathbb{R}^3$ is the gravity vector in the world frame; $\Delta t \in \mathbb{R}$ is the time elapsed between $i$ and $j$; $\Delta \tilde{v}_{ij} \in \mathbb{R}^3$ is the preintegrated velocity measurement; $\frac{\partial \Delta v}{\partial \mathbf{b}_g}, \frac{\partial \Delta v}{\partial \mathbf{b}_a} \in \mathbb{R}^{3 \times 3}$ are the Jacobians of the preintegrated velocity with respect to the gyroscope and accelerometer biases; and $\delta \mathbf{b}_{g,i}, \delta \mathbf{b}_{a,i} \in \mathbb{R}^3$ are the bias deviations from the linearization point at time $i$.

The positional error is represented as:
\begin{equation}
\label{eq:imu-factor-pos}
\begin{split}
\mathbf{e}_{\Delta p} ={}& R_i^\top \left( p_j - p_i - v_i\Delta t - \frac{1}{2}\mathbf{g}\Delta t^2 \right) \\
&- \left( \Delta \tilde{p}_{ij} + \frac{\partial \Delta p}{\partial \mathbf{b}_g} \delta \mathbf{b}_{g,i} + \frac{\partial \Delta p}{\partial \mathbf{b}_a} \delta \mathbf{b}_{a,i} \right),
\end{split}
\end{equation}
where $p_i, p_j \in \mathbb{R}^3$ are the positions of the link's CoM frame in the world frame at times $i$ and $j$; $\Delta \tilde{p}_{ij} \in \mathbb{R}^3$ is the preintegrated position measurement; and $\frac{\partial \Delta p}{\partial \mathbf{b}_g}, \frac{\partial \Delta p}{\partial \mathbf{b}_a} \in \mathbb{R}^{3 \times 3}$ are the Jacobians of the preintegrated position with respect to the gyroscope and accelerometer biases.

Lastly, the bias evolution error is:
\begin{equation}
\label{imu-factor-bias-evolution}
\mathbf{e}_{\Delta b} = \begin{bmatrix} \mathbf{b}_{g,j} - \mathbf{b}_{g,i} \\ \mathbf{b}_{a,j} - \mathbf{b}_{a,i} \end{bmatrix}.
\end{equation}
The bias evolution is modeled using a random zero-mean Gaussian walk. IMU preintegration \cite{gtsam-imu-3-preintegration} is utilized to consolidate hundreds of high-frequency IMU measurements into a single relative geometric constraint between two lower-frequency pose variables in the factor graph.

The standard combined IMU factor provided by GTSAM expects linear velocities in the world frame $v_i, v_j$ as an input. The proposed factor graph represents the velocity state using body twists $\xi_i, \xi_j \in \mathbb{R}^6$. Consequently, in our implementation the factor is modified to internally convert the body twist variables into linear spatial velocities. Let the body twist $\xi$ at time $i$ be partitioned into angular $\boldsymbol{\omega}_i$ and linear $\mathbf{v}_i^b$ components $\xi_i = [\boldsymbol{\omega}_i^\top, \mathbf{v}_i^{b\top}]^\top$, where $\mathbf{v}_i^b \in \mathbb{R}^3$ is the linear velocity expressed in the link's CoM frame. The linear spatial velocity $v_i$ is then derived by rotating the linear body velocity into the world frame:
\begin{equation}
\label{eq:twist-to-spatial-vel}
v_i = R_i \mathbf{v}_i^b.
\end{equation}
This transformation is applied to the body twist variables at both time slices $i$ and $j$ prior to evaluating the velocity error \eqref{eq:imu-factor-vel} and the subsequent positional error \eqref{eq:imu-factor-pos}.

\subsection{Contact}
\noindent
Contact is crucial in proprioceptive state estimation. Feet making contact with the ground serves as a static landmark that allows drift correction in the world frame $\mathcal{W}$. In our estimator we assume that the ground is flat and use three factors to model contact: \emph{height constraint factor}, \emph{zero velocity update (ZUPT) factor} and \emph{between pose factor}. 

The \emph{Height Constraint Factor} is a one-way factor that enforces a contact point of a foot to maintain a fixed height ($z=0$) in the world frame during the stance phase:
\begin{equation}
\label{eq:contact-height-factor}
\mathbf{e}_{h} = \mathbf{e}_3^\top \left( R_f {}^{f}p_c + p_f \right),
\end{equation}
where $\mathbf{e}_3 = [0, 0, 1]^\top$ is the unit vector aligned with the world $z$-axis; $R_f \in SO(3)$ and $p_f \in \mathbb{R}^3$ are the rotation matrix and translation vector of the foot link in the world frame, respectively; and ${}^{f}p_c \in \mathbb{R}^3$ is the constant position of the contact point expressed in the local foot link frame.

The \emph{ZUPT factor} is a unary factor that enforces the linear spatial velocity of the foot in contact to remain zero during the stance phase: 
\begin{equation}
\label{eq:contact-zupt}
\mathbf{e}_{\text{zupt}} = v_f,
\end{equation}
where $v_f \in \mathbb{R}^3$ is the linear spatial velocity of the foot link's CoM frame expressed in the world frame.

A \emph{BetweenFactor} is a two-way factor connecting contact foot pose variables at consecutive timesteps, enforcing the constraint that the foot remains stationary relative to the world frame throughout the duration of the stance phase:
\begin{equation}
\label{eq:contact-between-pose}
\mathbf{e}_{\text{pose}} = \log \left( \left( {}^{w}T_{f,i} \right)^{-1} {}^{w}T_{f,j} \right)^{\vee},
\end{equation}
where ${}^{w}T_{f,i}, {}^{w}T_{f,j} \in SE(3)$ are the poses of the foot link in the world frame at consecutive timesteps $i$ and $j$; $\log: SE(3) \to \mathfrak{se}(3)$ is the logarithmic map onto the Lie algebra; and $\vee: \mathfrak{se}(3) \to \mathbb{R}^6$ is the vee operator mapping the manifold error into the vector tangent space.

However, stationary feet assumptions frequently fail due to foot slippage. We address this in a probabilistic manner, by dynamically adjusting the covariance matrices of the contact factors based on vertical force sensor telemetry. 

The covariance $\Sigma$ for these contact factors is modulated continuously as an exponential decay function of the measured normal force:
\begin{equation}
\label{eq:contact-covariance}
\Sigma(F_z) = \Sigma_{\text{lo}} + (\Sigma_{\text{hi}} - \Sigma_{\text{lo}}) \exp\left(-\frac{F_z}{\alpha}\right),
\end{equation}
where $\Sigma_{\text{lo}}$ is the chosen strict covariance bound applied during firm contact, $\Sigma_{\text{hi}}$ is the chosen relaxed covariance bound permitting kinematic slip, $F_z$ is the measured total normal force at the foot, and $\alpha$ is a tunable scaling constant. 

The normal force $F_z$ is computed as the scalar sum of the four 1D force sensors on the respective foot link. For unary constraints like the ZUPT factor at timestep $i$, $F_z = F_{z,i}$. For the binary between pose factors connecting states at consecutive timesteps $i$ and $j$, the modulating force is chosen conservatively as $F_z = \min(F_{z,i}, F_{z,j})$.

\subsection{Graph Topology}

\begin{figure*}[!t]
\vspace*{3mm} 
\centering
\includegraphics[width=\textwidth]{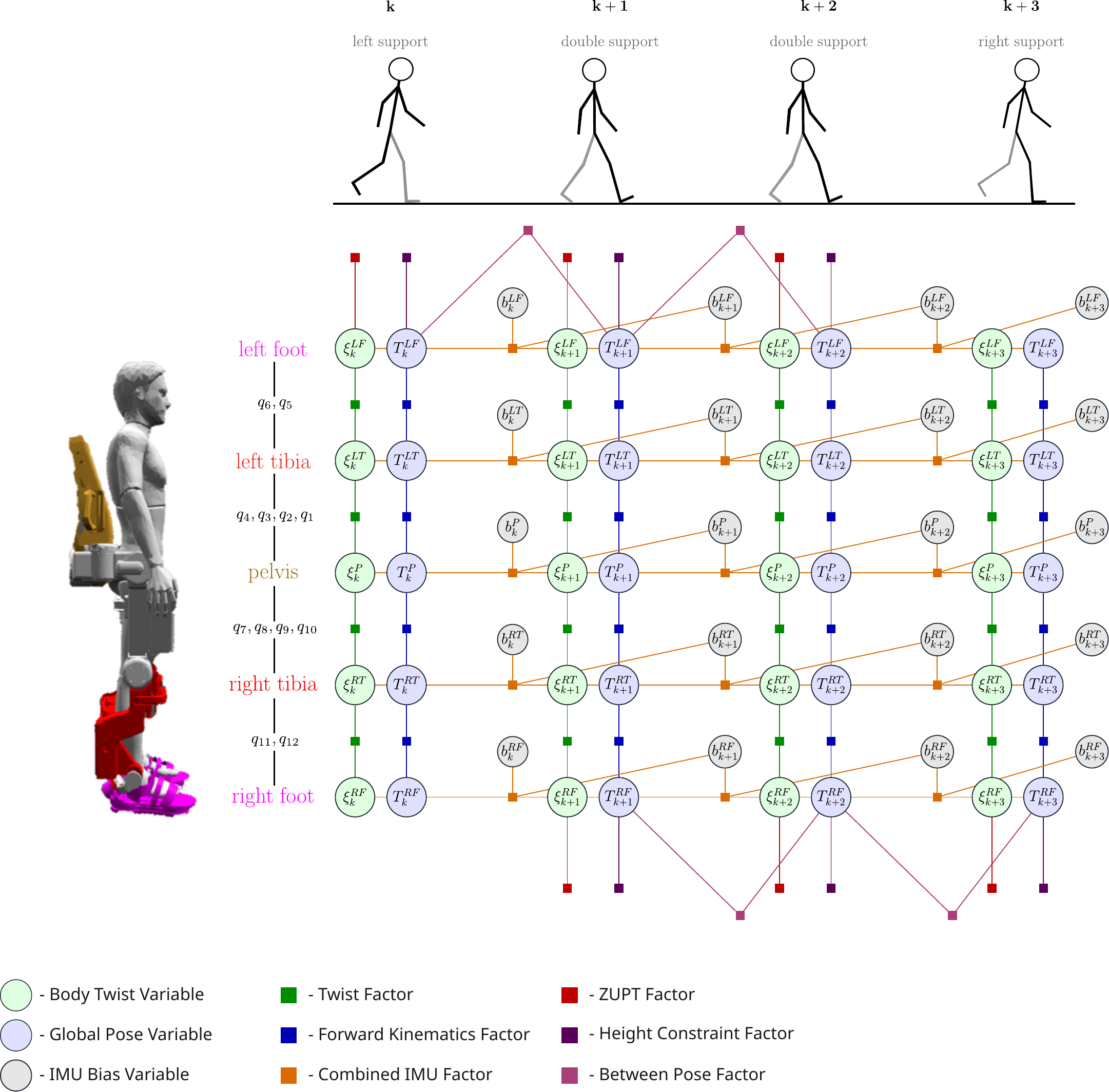}
\caption{Factor graph during walking. Four consecutive slices of the factor graph are shown: left support, double support, double support, right support. The contact factors are added to foot in contact during corresponding support stance. On the left $q_i$ represent which joint measurements are utilized by the corresponding kinematic factors. Joint numbering is the same as in Fig. \ref{fig:exo_frames}.}
\label{fig:graph-walking}
\end{figure*}

\noindent
Our factor graph is divided into slices, where each slice contains unknown pose variables $T_k$, body twists $\xi_k$, and IMU bias $b_k$ variables, with each associated with one particular timestep $k$. Kinematic factors, ZUPT factors and height constraint factors constrain variables within one slice, while combined IMU factors and between pose contact factors connect variables across consecutive slices. The topology of the factor graph is illustrated in Fig \ref{fig:graph-walking}.

Each slice includes kinematic variables associated with 5 links: pelvis, left/right tibia, left/right foot. We restrict our consideration to these links because this is where the IMU sensors are mounted. Foot links additionally have the force sensors and are the links in contact. Explicitly, each specified link contains the variables: 
\begin{itemize}
    \item $T_k \in SE(3)$ - pose of the link's CoM frame in the world frame at timestep k. 
    \item $\xi_k \in \mathbb{R}^6$ - twist of the link in link's CoM frame at timestep k.
\end{itemize}

Frequency of the graph is fixed, joint encoders measurements are linearly interpolated to fit the desired frequency, while IMU measurements are preintegrated. 

Proper initialization is required to establish the global reference frame and seed the estimator. At $t=0$, unary prior factors with strict covariances are applied to the initial pelvis pose to anchor the graph to the global reference frame. 

Force sensors on the feet are used to detect contact. Only when the overall force on a foot exceeds certain threshold, that foot is considered to be in contact. When a foot is in contact, contact factors are added to the corresponding foot link's variable, which is illustrated on Fig. \ref{fig:graph-walking}.

\section{Experiments}
\label{sec:experiment}
\noindent
We demonstrate the performance of our state estimator during dynamic walking of the Atalante exoskeleton with a human user. The estimated system state is benchmarked against a motion capture (mocap) ground truth and compared with EKF approaches and the MOVIE observer from \cite{MOVIE}. Additionally, we perform an ablation study to isolate the impact of our probabilistic kinematic and contact formulations.

\subsection{Experiment Setup}
\noindent
As detailed in Section \ref{sec:system-model}, the system uses 5 IMUs mounted on the pelvis, left/right tibia, and left/right feet, alongside four 1D force sensors per foot. All sensor telemetry is acquired at 1000 Hz. The factor graph optimizes the state trajectory at a frequency of 10 Hz. Consequently, the mocap ground truth data is downsampled and linearly interpolated to align with the 10 Hz estimation frequency for direct error evaluation.

To ensure accurate benchmarking, the global reference frames of the estimator and the motion capture system are pre-aligned using a prior factor on the pelvis link. For all experiments, static alignment biases between the motion capture ground truth and the IMU outputs are identified during a preliminary, one-time, stationary initialization phase. Furthermore, while the Combined IMU Factor dynamically estimates gyroscope additive bias, empirical observations indicate this bias remains highly stable over extended periods. 

Nonlinear factor graph optimization requires a robust initial estimate to ensure rapid convergence and avoid local minima. We employ a kinematics-driven sequential initialization strategy to compute the initial values for the solver. At $t=0$, the graph is anchored to the global reference frame by applying tight unary prior factors to the pelvis link, explicitly setting its pose to the motion capture ground truth and its spatial twist to zero to enforce a stationary initial state. The initial poses and twists of the remaining links in the kinematic chain (tibia and feet) are then deterministically populated by propagating the pelvis state via the system's forward kinematics and the joint measurements. For subsequent slices ($t>0$), the initial values for the optimizer are generated recursively. The pelvis pose and twist at timestep $k+1$ are predicted by propagating the pelvis state from timestep $k$ using the preintegrated inertial measurements from the pelvis IMU. Given this predicted pelvis state, the poses and twists of all links at $k+1$ are again computed directly via forward kinematics utilizing the joint encoder readings. For this experiments, the factor graph is constructed and optimized as an offline batch sequence.
\begin{figure*}[!t]
\vspace*{3mm} 
\centering
\includegraphics[width=0.85\textwidth]{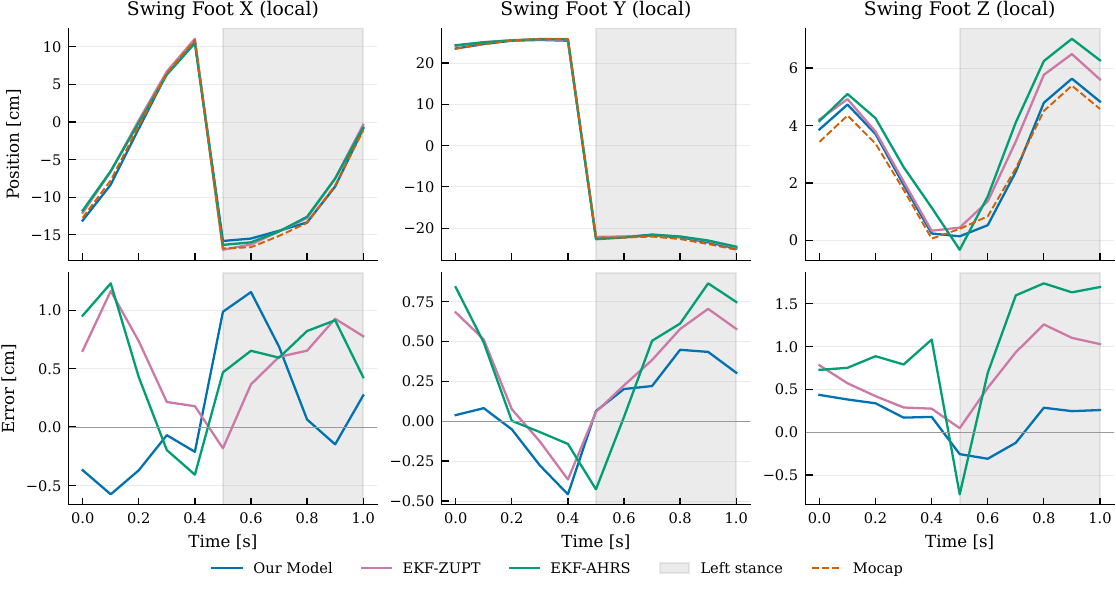}
\caption{Swing foot trajectory estimation in the local frame $\mathcal{L}$ during dynamic walking. The top row illustrates the estimated position against the motion capture ground truth across the sagittal (X), lateral (Y), and vertical (Z) axes. The bottom row displays the corresponding estimation error. The shaded area indicates when left foot is in contact.}
\label{fig:swing_foot_trajectory}
\end{figure*}

\begin{figure*}[!t]
\centering
\includegraphics[width=0.85\textwidth]{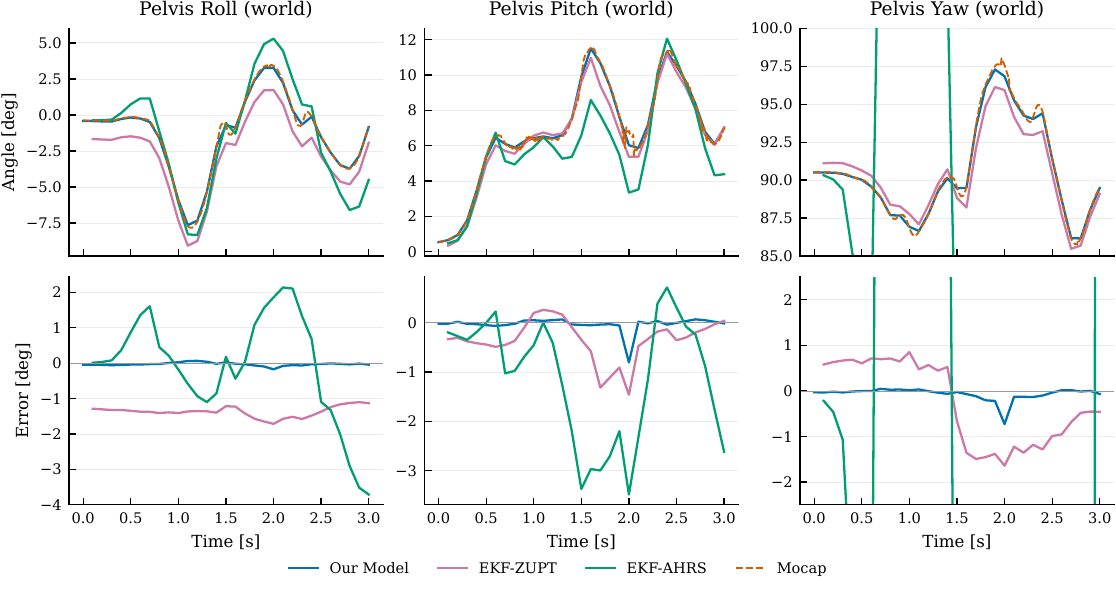}
\caption{Pelvis orientation estimation in the world frame. The top row illustrates the estimated orientation against the motion capture ground truth across the roll, pitch and yaw. The bottom row displays the corresponding estimation error.}
\label{fig:pelvis-world-orienation}
\end{figure*}

\subsection{EKF Baseline}
\noindent
Our primary baseline is an EKF with 15 states: pelvis-IMU attitude, position, velocity, accelerometer bias, and gyroscope bias. The pelvis IMU propagates the state at the IMU rate and measured joint angles and velocities provide rigid forward kinematics with no deformation modeling. The four remaining IMUs are treated as observations of a single rigid body chain.

The first baseline, denoted \emph{EKF-ZUPT}, corrects the state at every 10~Hz for each force-detected support foot by observing zero spatial velocity of the foot and the ground height $z=0$. The rest of the IMUs act as rigid-chain gyroscope-consistency measurements. By relying on kinematic zero-velocity constraints, this formulation separates linear acceleration from gravity.

The second baseline, denoted as \emph{EKF-AHRS}, replaces the ZUPT contact update with accelerometer-as-gravity corrections for the pelvis, and utilizes the rest of the IMUs to match their accelerometers to the predicted gravity direction from the robot's forward kinematics. This filter assumes that the sensor's linear acceleration is negligible on average and therefore treats the accelerometer as a gravity sensor.

Both filters do not model deformations and do not use force-dependent covariance schedule.

\subsection{Dynamic Walking}
\noindent
In this experiment, the exoskeleton walks forward in a straight trajectory for 15 seconds. We use an offline batch and Levenberg–Marquardt solver. For the 15 seconds walk, the estimator builds 10Hz factor graph with 150 slices and converges in 7.8 seconds. The results are summarized in Table \ref{tab:results}. All errors are presented using Root Mean Square Error (RMSE).

Fig. \ref{fig:swing_foot_trajectory} illustrates the estimated swing foot trajectory in the local frame $\mathcal{L}$ during stance switching. The proposed factor graph estimator closely tracks the motion capture ground truth across all three spatial axes. The feet positional errors remain bounded throughout the walk, achieving RMSE of under 0.7 cm. The EKF methods provide an accurate translational feet estimate in X and Y axis, but drift more in the Z axis.

Fig. \ref{fig:pelvis-world-orienation} presents the estimated pelvis orientation in the world frame $\mathcal{W}$. The estimation of the pelvis roll and pelvis pitch achieve RMSE values of around $0.04^\circ$ and $0.15^\circ$, respectively. Notably, the global yaw estimate remains stable and bounded over the whole walking sequence ($0.249^\circ$). We assume this improvement is due to the ability to take advantage of force-based dynamic covariance scheduling during contact.

In contrast, the baseline EKF methods show increased roll and pitch RMSE values (lowest RMSE being $0.53^\circ$ and $0.40^\circ$, respectively) with very poor yaw estimates ($0.76^\circ$). In particular, the method that uses accelerometer readings as gravity (EKF-AHRS) has no way of observing yaw at all, resulting in a very high error of $80^\circ$. 

Overall, our model significantly outperforms naive EKF approaches and provides results comparable with state-of-the-art MOVIE observer. While MOVIE has better estimate of feet orientation in local frame, our approach allows an accurate global yaw estimation.  

\subsection{Ablation Study}
\noindent
Lastly, to quantify the impact of our proposed method, we conduct an ablation study. The RMSE results evaluated against the motion capture ground truth are summarized in Table \ref{tab:ablation}.

To validate the hypothesis that a factor graph can reliably take into account structural deformations via probabilistic absorption, we compare our model against a strict rigid-body factor graph that uses $\times10^3$ smaller kinematic covariances. The resulting estimates exhibit consistent degradation in orientation and foot accuracy: pelvis world-frame pitch RMSE increases from $0.153^\circ$ to $0.274^\circ$, global yaw RMSE increases from $0.249^\circ$ to $0.342^\circ$, and swing foot local position RMSE increases from $6.94$ to $11.27$ mm. 

Next, we remove the dynamic adjustment of contact covariances. In this variant of the factor graph, force sensors are still used to detect contact, but covariances on contact factors remain fixed. Fixed contact covariances enforce a stronger stationary-foot prior throughout stance, which slightly reduces translational drift: pelvis world-frame translational RMSE decreased from $22.02$mm to $19.86$mm. However, during low-load contact transitions the foot actually slips. The rigid contact model interprets this motion as a global-state error and compensates by rotating the estimated trajectory, producing a substantially larger yaw error: increasing from $0.249^\circ$ to $0.867^\circ$. The proposed force-weighted model therefore trades a small increase in translational RMSE for a more reliable orientation estimate, particularly in global yaw.

\begin{table*}[t]
\vspace*{3mm} 
\footnotesize
\centering
\setlength{\tabcolsep}{3pt} 
\caption{Root Mean Square Error (RMSE) Comparison with Baselines}
\label{tab:results}
\begin{tabular}{lcccccccc}
\toprule
\textbf{Model} & 
\makecell{\textbf{Pelvis roll,} \\ \textbf{local [$^\circ$]}} & 
\makecell{\textbf{Pelvis pitch,} \\ \textbf{local [$^\circ$]}} & 
\makecell{\textbf{Foot roll,} \\ \textbf{local [$^\circ$]}} & 
\makecell{\textbf{Foot pitch,} \\ \textbf{local [$^\circ$]}} & 
\makecell{\textbf{Pelvis yaw,} \\ \textbf{world [$^\circ$]}} & 
\makecell{\textbf{Pelvis pos,} \\ \textbf{local [mm]}} & 
\makecell{\textbf{Pelvis pos,} \\ \textbf{world [mm]}} & 
\makecell{\textbf{Swing foot pos,} \\ \textbf{local [mm]}} \\
\midrule

EKF --- AHRS & 
1.794 & 1.180 & 2.234 & 1.291 & 80.59 & 31.87 & 2891.7 & 14.88 \\[1ex]

EKF --- ZUPT & 
0.533 & 0.394 & 0.863 & 0.409 & 0.757 & 11.26 & 59.0 & 12.6 \\[1ex]

MOVIE$^{\mathrm{a}}$ observer \cite{MOVIE} & 
0.18 & 0.17 & \textbf{0.18} & \textbf{0.22} & N/A & N/A & N/A & 14.1 \\[1ex]

Ours (full model) & 
\textbf{0.042} & \textbf{0.153} & 0.21 & 0.32 & \textbf{0.249} & \textbf{7.34} & \textbf{22.02} & \textbf{6.94} \\[1ex]

\bottomrule

\multicolumn{9}{l}{\rule{0pt}{3ex}$^{\mathrm{a}}$The MOVIE observer only provides estimates in the local frame $\mathcal{L}$, as its formulation centers the yaw.}
\end{tabular}
\end{table*}

\begin{table*}[t]
\footnotesize
\centering
\setlength{\tabcolsep}{3pt} 
\caption{Ablation Study RMSE Results}
\label{tab:ablation}
\begin{tabular}{lcccccccc}
\toprule
\textbf{Variant} & 
\makecell{\textbf{Pelvis roll,} \\ \textbf{local [$^\circ$]}} & 
\makecell{\textbf{Pelvis pitch,} \\ \textbf{local [$^\circ$]}} & 
\makecell{\textbf{Foot roll,} \\ \textbf{local [$^\circ$]}} & 
\makecell{\textbf{Foot pitch,} \\ \textbf{local [$^\circ$]}} & 
\makecell{\textbf{Pelvis yaw,} \\ \textbf{world [$^\circ$]}} & 
\makecell{\textbf{Pelvis pos,} \\ \textbf{local [mm]}} & 
\makecell{\textbf{Pelvis pos,} \\ \textbf{world [mm]}} & 
\makecell{\textbf{Swing foot pos,} \\ \textbf{local [mm]}} \\
\midrule

Ours w/ tight kinematics & 
\makecell{0.107} & 0.274 &  \makecell{0.3} & 0.35 &  0.342 & 8.27 & 22.5 & 11.27 \\[1ex]

Ours w/o contact & 
\makecell{0.045} & 0.241 &  \makecell{0.415} & 0.493 &  0.867 & \textbf{7.08} & \textbf{19.86} & 7.20 \\[1ex]

Ours (full model) & 
\textbf{0.042} & \textbf{0.153} & \textbf{0.21} & \textbf{0.32} & \textbf{0.249} & 7.34 & 22.02 & \textbf{6.94} \\[1ex]

\bottomrule
\end{tabular}
\vspace{-4mm}
\end{table*}

\section{Conclusion}
\label{sec:conclusion}
\noindent

This paper presented a proprioceptive state estimation framework for the Atalante lower-limb exoskeleton using factor graphs. By probabilistically absorbing structural deformations and dynamically adjusting contact covariances based on real-time force sensor telemetry, the proposed smoother provides an accurate estimation for the pelvis and feet link in both world $\mathcal{W}$ and local $\mathcal{L}$ frames. Experimental validation against motion capture ground truth have demonstrated an advantage of temporal factor graph smoothing over naive EKF approaches. 

While achieving state-of-the-art performance, the current implementation is limited to an offline batch optimization, that takes $7.8$s to optimize $15$s walk, which is not optimal for real time deployment in the safety-critical system such as Atalante. At the same time, we didn't take advantage of modern incremental optimization strategies such as iSAM2 \cite{isam2}. We believe that with iSAM2, the proposed architecture will be able to satisfy low-latency requirements of modern exoskeleton systems. 

Besides real-time online performance, future work will extend this framework to incorporate full human-user body dynamics into the kinematic model. Furthermore, we intend to explore data-driven learning methods to optimize the dynamic covariance scheduling.

\section{Acknowledgments}
The EKF baseline was produced using GitHub Copilot Claude, model Claude Opus 5.


\balance
\bibliographystyle{IEEEtran}
\bibliography{references}

\newpage

\vfill

\end{document}